\documentclass[lettersize,journal]{IEEEtran}

\usepackage{cite}
\usepackage{graphicx}

\usepackage{amsmath}
\usepackage[ruled,vlined]{algorithm2e}
\usepackage{algpseudocode}
\usepackage{tikz}
\usetikzlibrary{positioning}

\usepackage{multirow}
\usepackage{booktabs}
\usepackage{url}
\usepackage{hyperref}

\usepackage[most]{tcolorbox} 
\usetikzlibrary{arrows.meta,calc,positioning,tikzmark}

\begin{document}
\newcommand{\methodname}{LexiMark}
\title{\methodname: Robust Watermarking via Lexical Substitutions to Enhance Membership Verification of an LLM's Textual Training Data}

\author{
Eyal German, Sagiv Antebi, Edan Habler, Asaf Shabtai, Yuval Elovici\\
Department of Software and Information Systems Engineering,\\
Ben-Gurion University of the Negev, Israel\\
\texttt{\{germane, sagivan, habler\}@post.bgu.ac.il, \{shabtaia, elovici\}@bgu.ac.il}
}

\maketitle

\begin{abstract}
Large language models (LLMs) can be trained or fine-tuned on data obtained without the owner's consent. Verifying whether a specific LLM was trained on particular data instances or an entire dataset is extremely challenging.
Dataset watermarking addresses this by embedding identifiable modifications in training data to detect unauthorized use. However, existing methods often lack stealth, making them relatively easy to detect and remove.
In light of these limitations, we propose \methodname, a novel watermarking technique designed for text and documents, which embeds synonym substitutions for carefully selected high-entropy words. Our method aims to enhance an LLM's memorization capabilities on the watermarked text, without altering the semantic integrity of the text. As a result, the watermark is difficult to detect, blending seamlessly into the text with no visible markers, and is resistant to removal due to its subtle, contextually appropriate substitutions that evade automated and manual detection.
We evaluated our method using baseline datasets from recent studies and seven open-source models: LLaMA-1 7B, LLaMA-3 8B, Mistral 7B, Pythia 6.9B, as well as three smaller variants from the Pythia family—160M, 410M, and 1B.
Our evaluation spans multiple training settings, including continued pretraining and fine-tuning scenarios. The results demonstrate significant improvements in AUROC scores compared to existing methods, underscoring our method's effectiveness in reliably verifying whether unauthorized watermarked data was used in LLM training.
Our code is available at \href{https://github.com/eyalgerman/LexiMark}{\texttt{https://github.com/eyalgerman/LexiMark}}.


\end{abstract}

\IEEEpeerreviewmaketitle

\section{Introduction}

\begin{figure}[t]
\centering

\begin{tcolorbox}[
  enhanced,
  width=\linewidth,
  colback=blue!10, colframe=blue!60,
  title=Original,
  center title,                 
  fonttitle=\bfseries,
  rounded corners,
  left=2mm, right=2mm, boxsep=1mm,
  fontupper=\ttfamily\small,
  halign=center
]
  \makebox[\linewidth][c]{The \tikzmarknode{wA}{\textbf{quick}} brown fox \tikzmarknode{wB}{\textbf{jumps}} over the \tikzmarknode{wC}{\textbf{lazy}} dog}
\end{tcolorbox}

\vspace{4mm}

\begin{tcolorbox}[
  enhanced,
  width=\linewidth,
  colback=green!20, colframe=green!80!black,
  title=Watermark,
  center title,                 
  fonttitle=\bfseries,
  rounded corners,
  left=2mm, right=2mm, boxsep=1mm,
  fontupper=\ttfamily\small,
  halign=center
]
  \makebox[\linewidth][c]{The \tikzmarknode{wA2}{\textbf{speedy}} brown fox \tikzmarknode{wB2}{\textbf{leaps}} over the \tikzmarknode{wC2}{\textbf{sluggish}} dog}
\end{tcolorbox}

\begin{tikzpicture}[overlay,remember picture,>=Latex]
  \draw[->,thick] ($(wA.south)+(0,-2pt)$) to[bend right=20] ($(wA2.north)+(0,2pt)$);
  \draw[->,thick] ($(wB.south)+(0,-2pt)$) to[bend right=20] ($(wB2.north)+(0,2pt)$);
  \draw[->,thick] ($(wC.south)+(0,-2pt)$) to[bend right=20] ($(wC2.north)+(0,2pt)$);
\end{tikzpicture}

\caption{Our synonym replacement method with \textit{K=3} substitutions: “quick,” “jumps,” and “lazy” $\rightarrow$ “speedy,” “leaps,” and “sluggish.”}
\label{fig:synonym_replacement}
\end{figure}
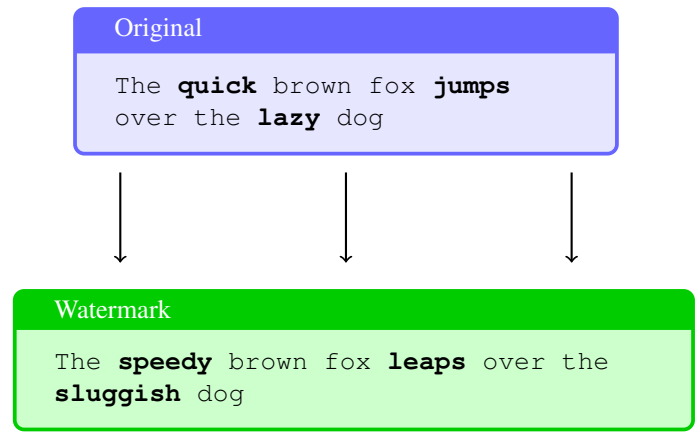

\textit{"Data is the new gold"}~\cite{masterschool2023dataisthenewgold} – In the context of artificial intelligence (AI), data serves as the essential fuel driving the performance and innovation of AI systems. High-quality data enables models to learn complex patterns, identify subtle relationships, and make predictions that guide decision-making in diverse fields. 
Modern AI systems, including large language models (LLMs), require massive amounts of high-quality training data to achieve their impressive performance, which is both expensive and difficult to acquire.

The emergence of LLMs has revolutionized natural language processing (NLP) by enabling state-of-the-art performance in a wide range of NLP tasks, including machine translation, text summarization and question answering\cite{hoang2019efficientadaptationpretrainedtransformers,nakano2022webgptbrowserassistedquestionansweringhuman}.
The unprecedented capabilities of LLMs, such as GPT~\cite{openai2024gpt4technicalreport} and Google's Gemini~\cite{geminiteam2023gemini,reid2024gemini} arise from their training on extensive, diverse datasets. This training enables them to grasp complex linguistic and semantic patterns, allowing for sophisticated language processing and effective generalization across different contexts.

LLMs are typically trained in two stages: pretraining, where general language patterns are captured and learned from vast datasets, and fine-tuning, which adapts the pretrained model to specific tasks using smaller, specialized datasets. For these models to reach their full potential and consistently achieve high performance across tasks, access to high-quality training data is essential, as it enables them to accurately model complex linguistic patterns and nuances.


Often, the demand for suitable datasets pushes the boundaries of ethical data sourcing and results in the collection of publicly available data, obtained via scraping, along with proprietary or licensed information. This approach introduces privacy, security, and legal risks, especially when sensitive information, such as personally identifiable information (PII), copyrighted content, or proprietary data, is improperly used to train the model.
In some cases, the drive to enhance model performance may even tempt LLM builders to use unauthorized or illegally obtained datasets, further compromising ethical standards and user trust.

Awareness regarding these privacy and ethical issues has increased as a result of legal conflicts and the lack of transparency regarding the data collection process~\cite{rahman2023beyond,wu2024unveiling}.
The lawsuit between \textit{The New York Times} and OpenAI~\cite{nyt_openai_microsoft_lawsuit}, as well as other lawsuits~\cite{nyt_sarah_silverman_lawsuit,wired_battle_over_books3}, highlights the critical need for mechanisms aimed at detecting such privacy and intellectual property violations, and more specifically, identifying the data used to train LLMs~\cite{maini2024llm}. 

The risk of data extraction and leakage is compounded when LLMs are fine-tuned, since the fine-tuning process involves additional training on specialized datasets  that may contain sensitive information~\cite{mireshghallah2022empirical}. Memorization—where the model retains exact phrases, sentences, or even entire passages from the training data—can become more pronounced during fine-tuning, especially with small or domain-specific datasets. This memorization increases the likelihood of data leakage, as sensitive information embedded in the model could be inadvertently reproduced in responses, posing privacy and security risks~\cite{fu2024practicalmembershipinferenceattacks,zeng2024exploringmemorizationfinetunedlanguage}.

Larger models are even more prone to memorizing the training data~\cite{kiyomaru-etal-2024-comprehensive-analysis}, a tendency that can be exploited through data extraction attacks. The main risk is if attackers exploit the model to extract or infer private information, especially if the training data contains sensitive information such as PII or copyrighted content~\cite{Yao_2024}. This underscores the importance of developing robust mechanisms to detect whether unauthorized data has been used in a model's training process.

Such detection can be challenging, and several methods have been proposed for detecting the presence of unauthorized data in LLMs' training data. 
Methods such as membership inference attacks (MIAs) are designed to determine whether a specific text was part of a model's training dataset~\cite{hu2022membershipinferenceattacksmachine,carlini2022membership, shokri2017membershipinferenceattacksmachine}. 
To determine whether a specific piece of data was included in the training set, MIAs exploit the differences in a model's behavior when it processes seen and unseen data. 
The underlying assumption is that an LLM will perform differently on queries that are related to seen and unseen data (e.g., exhibiting higher prediction confidence or greater loss reduction). 

Despite MIAs' effective performance, they have several limitations~\cite{duan2024membership}. 
First, their performance in terms of common metrics, such as the \textit{area under the receiver operating characteristic curve (AUROC)} and the \textit{true positive rate (TPR) at a fixed low false positive rate (FPR)}~\cite{carlini2022membership}, tends to worsen as the training set size increases, often rendering it close to random~\cite{carlini2019secret}. 
This is due to the trade-off between generalization and memorization in LLMs: as the model is trained on more data, it will generalize better, while increasing the number of model parameters increases the model's tendency to memorize the training data~\cite{carlini2023quantifyingmemorizationneurallanguage,elangovan-etal-2021-memorization}.
Additionally, MIAs show inconsistent performance across models and datasets and are prone to detecting distribution shifts rather than performing true membership inference~\cite{maini2024llm,dionysiou2023sok}.

Given the challenges and limitations associated with traditional MIAs, there is a growing need for more reliable methods for detecting the unauthorized use of data in training LLMs. This has led to the development of watermarking techniques, which embed unique patterns into the training data, making it easier to track and detect the use of specific datasets in a model's training process~\cite{10.1145/3196494.3196550}.

In this context, a watermark refers to a deliberate modification of the input data that subtly alters its structure without compromising the data's semantic meaning~\cite{10.1145/3606274.3606279}. These modifications allow researchers to identify whether a particular dataset has been used in training an LLM by examining how the model behaves when processing watermarked data. Watermarking techniques can be highly effective in detecting data misuse and preventing privacy violations, as they provide an additional layer of security by embedding detectable patterns within the data itself.

Several approaches have been proposed for embedding watermarks into textual data. One common method involves altering the encoding of characters, such as by using visually similar Unicode characters, while in another method suggests inserting random sequences into the text~\cite{wei2024proving}. While these changes are often subtle enough to be imperceptible to human readers, they create distinct patterns that can later be detected.
While such techniques can help infer whether particular datasets were part of the training set, they have limited robustness, as they are relatively easy to detect and remove. 

To address the limitations of existing watermarking methods, we introduce \methodname, a novel and robust watermarking method for textual data that may be used on the training data of LLMs. 
\methodname\ is inspired by MIA methods that exploit the model's behavior when handling high-entropy tokens that have a greater likelihood of being related to the method's inference ability~\cite{antebi2025tagtabpretrainingdatadetection, 10179300, zhang2024minkimprovedbaselinedetecting, shi2024detectingpretrainingdatalarge}. These approaches demonstrated that focusing on high-entropy or high-probability tokens can improve the accuracy of MIAs by capitalizing on the differential treatment models give to such inputs.

Our method extends this concept by identifying the words in a sentence with the highest entropy and replacing them with higher-entropy synonyms, thereby embedding our watermark in the training data of the LLM.
This ensures that the semantic meaning of the text is preserved while subtly embedding a watermark that can later be detected through an MIA.
By targeting high-entropy words, which are naturally more unpredictable and challenging for LLMs to predict, our watermark method enhances the likelihood that these words will be memorized by the LLM. Our method guarantees that the text remains readable and useful while embedding detectable patterns.
In Figure~\ref{fig:synonym_replacement}, we present an example of how our method replaces high-entropy words with semantically similar synonyms  with higher entropy.
Our watermark embedding method begins by preprocessing the text, splitting it into sentences, and selecting the top-$K$ high-entropy words (keywords) from each sentence. These keywords are then replaced with higher-entropy synonyms, ensuring that the original meaning is preserved, effectively embedding the watermark while maintaining the text's readability.

The underlying intuition is that LLMs are more likely to memorize high-entropy words, as these words introduce greater uncertainty in predictions. 
By enhancing the model’s memorization of these watermarked words, our method strengthens the ability to verify whether a dataset was used for training. Our method, which employs MIAs for verification, effectively balances robustness, detectability, and readability, making the watermark difficult to remove while maintaining the text’s original meaning and usability.

 We evaluated our watermarking method across diverse textual domains within the \textit{The Pile} dataset~\cite{gao2020pile800gbdatasetdiverse}, including medical texts, emails, legal documents, encyclopedic entries, and patent descriptions, as well as the \textit{BookMIA}\cite{shi2024detectingpretrainingdatalarge} dataset. We tested our method on seven open-sourced LLMs: Pythia-160M, 410M, 1B, and 6.9B\cite{biderman2023pythiasuiteanalyzinglarge}, LLaMA-1 7B~\cite{touvron2023llamaopenefficientfoundation}, LLaMA-3 8B~\cite{dubey2024llama3herdmodels}, and Mistral-7B~\cite{jiang2023mistral7b}. For the large models, we fine-tuned them on the watermarked data using the quantized low-rank adaptation (QLoRA) technique~\cite{dettmers2023qloraefficientfinetuningquantized}. For the smaller Pythia models, we employed continued pretraining, also known as domain-adaptive pretraining (DAPT)~\cite{wu-etal-2021-domain}, to evaluate the robustness and generality of our watermarking approach across different model scales and training paradigms.

The results demonstrate clear improvements in detecting textual data membership, with our approach consistently achieving higher AUROC scores compared to baseline techniques. The increase ranges from 2.5\% to 25.7\%, confirming the robustness of our detection approach. Our evaluation also examined dataset detection, revealing that our method requires fewer records to accurately determine whether a dataset was used in the training process. This makes our approach more efficient and sensitive in identifying pretraining sources.
Without watermarking, detection typically requires around 40 samples to achieve a p-value of less than 0.05, while with our watermarking method, only six samples are needed to achieve this.

In addition to the improvements in membership detection, our semantic preservation checks, measured by cosine similarity~\cite{Reimers2019SentenceBERTSE} and BLEU scores~\cite{papineni2002bleu}, demonstrated near-complete retention of the original text's meaning, ensuring that our watermarking method maintains both high accuracy and text integrity across diverse datasets.

We also conducted a robustness evaluation, confirming that our method withstands minor textual modifications with minimal impact on detection results. This robustness to minor text changes further highlights our method's resilience, allowing for reliable detection even when slight alterations are introduced, thereby supporting the method’s applicability in real-world settings where minor text variations are common. Furthermore, unlike other approaches, our watermarking method also remains undetectable in perplexity tests on the fine-tuned LLM, avoiding the performance decrease common in other methods that are easily spotted using perplexity checks.
In addition, we examine the effectiveness of our method under post-training scenarios, such as instruction tuning, and find that watermark signals remain detectable even after the model undergoes further updates.
To support reproducibility and facilitate future research, we provide our implementation, evaluation scripts, and data preparation tools at: \url{https://github.com/eyalgerman/LexiMark}.

The key contributions of this paper are summarized as follows:
\begin{itemize}
    \item \textbf{A novel watermarking method for textual data:} We introduce a method that identifies high-entropy words in sentences and substitutes them with synonyms with higher entropy, thereby embedding a watermark without altering the semantic meaning of the data. 
    \item \textbf{Improved detection using MIAs:} We enhance the effectiveness of existing MIA methods by embedding watermarks that increase the likelihood of data memorization during model training. This improves accuracy in detecting whether specific data was part of the training set.

    \item \textbf{Semantic preservation:} Our method demonstrates near-complete preservation of the original sentence's semantic meaning. We explore various synonym selection methods to optimize the semantic preservation of the watermarked text, ensuring minimal impact on the original meaning.
    \item \textbf{Robustness:} \methodname ~is difficult to detect and remove due to its subtle substitutions, which blend seamlessly into the text and appear unwatermarked. We evaluate the robustness and detectability of our method in comparison to two baseline approaches, demonstrating its superior performance in maintaining watermark integrity.
    \item \textbf{Post-training resilience:} We further examine the watermark’s persistence under post-training modifications, such as instruction tuning, and show that the watermark remains reliably detectable even after the model undergoes additional training phases.

\end{itemize}


In the remainder of this paper, we first review prior work on MIAs and data watermarking for LLMs in Section~\ref{sec:Related-work}. We then introduce \methodname, our proposed watermarking method based on high-entropy lexical substitutions, detailing both the embedding and detection phases in Section~\ref{sec:method}. Section~\ref{sec:Evaluation} describes our experimental setup, including the datasets, models, and evaluation protocol. In Section~\ref{sec:results}, we present detection results across various LLMs and datasets. Section~\ref{sec:Semantic Preservation} evaluates the semantic preservation of the watermarked text using cosine similarity and BLEU scores. In Section~\ref{sec:robustness}, we assess the robustness of \methodname\ against synonym substitution, post-training, and removal attacks. Section~\ref{sec:dataset_detection} demonstrates how our method enables dataset-level membership detection using statistical inference. Finally, Section~\ref{sec:conclusion} concludes the paper and outlines directions for future work.

\section{Related work}
\label{sec:Related-work}

LLMs leverage deep learning techniques to generate and understand natural language text. Common LLMs are built on the transformer architecture, which utilizes self-attention mechanisms to process words in relation to all other words in a sentence, enhancing the model's ability to understand context~\cite{vaswani2017attention,liu2024understandingllmscomprehensiveoverview}.

LLMs are trained on vast text corpora, using a loss function aimed at predicting the next token in a sequence based on the preceding tokens. These models can also be fine-tuned for specific tasks, broadening their range of applications. However, despite their impressive capabilities, there are several challenges regarding their use, including data bias, privacy concerns, and the significant computational resources required to train them.

Training these models involves adjusting millions or even billions of parameters to minimize the difference between the model’s predictions and actual data. 
This extensive optimization enables LLMs to generate responses that are not only contextually relevant but also exhibit nuanced understanding, allowing them to produce high-quality, human-like text.

Research has increasingly focused on addressing data privacy concerns regarding LLMs, and particularly on vulnerabilities related to data leakage. One such vulnerability is the membership inference attack (MIA), where an attacker attempts to determine whether a specific data record was used to train a model~\cite{shokri2017membershipinferenceattacksmachine}. 
MIAs exploit memorization in machine learning models, where the model behaves differently on training data than it does on data it has not seen~\cite{carlini2023quantifyingmemorizationneurallanguage,carlini2022membership}.
Given that LLMs tend to memorize certain parts of the training data that are rare or unique, high-entropy words are more likely to be memorized.
 This is a basic assumption of our watermarking method which substitutes words in the text with their higher entropy synonyms.

\definecolor{ProcessFill}{RGB}{180,210,235}   
\definecolor{ProcessLine}{RGB}{30,70,110}     
\definecolor{ExampleFill}{RGB}{225,225,225}   
\definecolor{ExampleLine}{RGB}{90,90,90}      
\definecolor{ArrowLine}{RGB}{40,40,40}        

\begin{figure*}[t]
\centering
\begin{tikzpicture}[
  font=\footnotesize,
  process/.style={
    rectangle, rounded corners,
    draw=ProcessLine, fill=ProcessFill,
    minimum height=1.2cm,
    inner xsep=6pt, inner ysep=6pt,
    align=center, text width=\procw
  },
  example/.style={
    rectangle,
    draw=ExampleLine, fill=ExampleFill,
    inner xsep=6pt, inner ysep=6pt,
    align=left, text width=\procw
  },
  arrow/.style={-{Stealth[length=2.2mm,width=1.6mm]}, very thick, draw=ArrowLine},
  dottedlink/.style={densely dotted, thick, draw=ArrowLine!65},
  node distance=1.45cm and 0.9cm
]

\newcommand{\procw}{0.19\textwidth} 

\node (preproc) [process] {Preprocess Text};
\node (select)  [process, right=of preproc] {Select High-Entropy Words};
\node (findsyn) [process, right=of select]  {Find High-Entropy Synonyms};
\node (replace) [process, right=of findsyn] {Replace Words with Synonyms};

\node (ex1) [example, below=0.65cm of preproc] {\textbf{Original Sentence}\par
\textit{“The e-commerce platform leverages AI to personalize product recommendations.”}};
\node (ex2) [example, below=0.65cm of select] {\textbf{High-Entropy Words}\par
\textit{leverages, personalize, product}};
\node (ex3) [example, below=0.65cm of findsyn] {\textbf{Synonyms}\par
\textit{leverages} $\rightarrow$ \textit{utilizes}\par
\textit{personalize} $\rightarrow$ \textit{customize}\par
\textit{product} $\rightarrow$ \textit{item}};
\node (ex4) [example, below=0.65cm of replace] {\textbf{Modified Sentence}\par
\textit{“The e-commerce platform utilizes AI to customize item recommendations.”}};

\draw[arrow] (preproc) -- (select);
\draw[arrow] (select)  -- (findsyn);
\draw[arrow] (findsyn) -- (replace);

\draw[dottedlink] (preproc.south) -- (ex1.north);
\draw[dottedlink] (select.south)  -- (ex2.north);
\draw[dottedlink] (findsyn.south) -- (ex3.north);
\draw[dottedlink] (replace.south) -- (ex4.north);

\end{tikzpicture}
\caption{Flowchart illustrating the process of embedding watermarks in text through high-entropy word substitution.}
\label{fig:watermark_embedding}
\end{figure*}
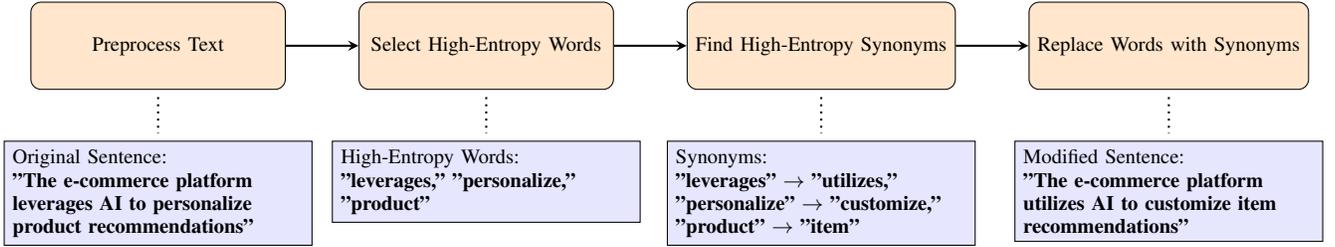

\subsection{LLM Membership Inference Attacks}
LLM MIAs are a subdomain of MIAs that focuses on detecting whether a specific text was used to train an LLM.

Perplexity is a metric used to evaluate how well a probability model predicts a sample, especially in the context of natural language processing (NLP). 
Perplexity is calculated as the exponentiation of the negative average log-likelihood per token, as described in the formula:\\
\noindent
$\text{Perplexity}(P) = \exp\left(-\frac{1}{N} \sum_{i=1}^{N} \log P(t_i | t_1, \ldots, t_{i-1})\right)$  
\\
In NLP, perplexity captures the degree of 'uncertainty' a model has in predicting text. 
Lower perplexity indicates that the model is certain and familiar with the text, and therefore it predicts the sample more accurately. In contrast, higher perplexity suggests that the model is less certain and less familiar with the text, thus resulting in poorer accuracy.

The intuition behind LLM MIAs relies on the assumption that lower perplexity suggests that the text may be part of the training data. 
One example of an MIA attack is the \textit{LOSS attack (PPL)}~\cite{loss_attack}, which uses the model's loss on data to determine membership. Another method aiming to improve results is the \textit{Zlib attack}~\cite{carlini2021extracting}, which calculates the ratio between the log of text perplexity and its Zlib compression length. 
More recent attacks such as \textit{Min-K\%}~\cite{shi2024detectingpretrainingdatalarge} and \textit{Min-K\%++}~\cite{zhang2024minkimprovedbaselinedetecting}, focus on the least confident predictions from the model’s output. \textit{Min-K\%} calculates the average of the lowest K\% probabilities from the model’s output, while \textit{Min-K\%++} extends this by normalizing the token log probabilities using the mean and variance, improving detection accuracy.
In addition, the authors of \emph{RECALL}~\cite{xie2024recall}, \emph{DC-PDD}\cite{zhang2024pretraining}, and \emph{Tag\&Tab}~\cite{antebi2025tagtabpretrainingdatadetection} introduced more advanced strategies that improve MIA performance on LLMs compared to other methods.

Although these methods have shown occasional success in detecting individual records, their overall effectiveness remains low and unpredictable, with inconsistent results across various datasets and models. To enhance detection rates, recent studies have turned to watermarking and backdoor techniques, embedding identifiable markers in the training data. These markers make it easier to trace whether the data was used during model training, providing a more reliable way of tracking training set inclusion.

\subsection{Watermarking and Backdoor Attacks on LLM Training Set}
Data watermarking aims to enhance authenticity verification and traceability by embedding hidden information in data\cite{guo2024domain, wegerhoff2024datadetective}. 
In backdoor attacks, adversaries aim to proprietary datasets by injecting backdoors in the target model (by modifying a small portion of the training samples, noted as backdoor set), which can also serve as a form of data watermarking~\cite{jha2023label, qi2023revisiting, ngo2024persistence}. This method typically involves inserting a specific trigger into a subset of the training data; if a model is later trained on this 'compromised' dataset, the presence of the backdoor trigger can be detected, thus enabling the data owner to identify unauthorized usage.

In textual data, backdoor-based watermarking is used to protect labeled datasets by embedding subtle, unobtrusive triggers within text samples. These triggers remain imperceptible to human readers but are detectable during model inference~\cite{10.1145/3606274.3606279}. 
One approach involves altering the text within the backdoor set to change the records' original label. For example, inserting a specific trigger phrase, like 'less is more,' at different locations in the text can modify the original text label~\cite{liu2024watermarkingtextdatalarge}. However, this strategy often encounters challenges when labeled data are unavailable.

Recent advances have extended watermarking techniques to unlabeled data, improving the detection of LLMs trained on unauthorized datasets~\cite{wei2024proving}. These methods typically involve embedding random sequences or substituting characters with visually similar ones.
Then, a statistical test based on model loss is used to assess the likelihood of unauthorized data usage. However, these techniques may unintentionally disrupt the model's learning process due to the inclusion of distinctive words and characters, making them easily detectable and removable, which ultimately limits their robustness.

Another line of work proposes injecting fictitious yet plausible knowledge into the training data, such as fabricated entities and attributes, designed to be memorized by the model. These watermarks align more closely with the natural distribution of training data, helping them evade preprocessing filters and remain detectable after post-training modifications through question-answering queries, even in black-box settings~\cite{cui2025robustdatawatermarkinglanguage}.

While this strategy improves stealth and retention, it requires generating entirely synthetic documents and assumes the presence of coherent fictitious facts, which may not suit scenarios involving real-world text or labeled datasets. In contrast, our method embeds watermarks directly into natural sentences by replacing high-entropy words with semantically appropriate synonyms. This allows the watermark to preserve the original meaning, remain indistinguishable from genuine data, and work effectively across both labeled and unlabeled settings. Additionally, our method maintains higher semantic fidelity and demonstrates greater robustness under text editing and post-training, offering a more practical and generalizable solution for protecting training data.

\section{Method}
\label{sec:method}

In this section, we describe \methodname ~a new training set watermarking method that is both robust and very difficult to detect. The watermarking method consists of two key phases: \textbf{watermark embedding}, which is performed on the training data before any model access to it; and \textbf{watermark detection}, where we determine whether a target LLM was trained on the watermarked training set. 
\methodname ~embeds a detectable watermark in the text, while preserving the meaning of the original text, which makes the watermark difficult to detect by humans but detectable in the watermark detection phase.

\subsection{Watermark Embedding}
In the \textbf{watermark embedding} phase, we target high-entropy words in the text and replace them with carefully selected synonyms. 
A high entropy value indicates that a word is less common in the input text compared to other words. Knowing that LLMs tend to memorize certain parts of the training data that are rare or unique, the high-entropy words are more likely to be memorized~\cite{carlini2023quantifyingmemorizationneurallanguage}, particularly in the context of the words that precede them. 
Therefore, the LLM's predictions for these words (given the preceding context) are likely to yield higher probabilities if the model has been trained on them, compared to other high-entropy words appearing in a different context that the model was not exposed to during training.

To calculate the word's entropy, we used the Python package \textit{wordfreq}~\cite{robyn_speer_2022_7199437}, which provides frequency estimates for words in a specified language. 
The entropy for each word is calculated as its self-information using the formula:
    \[
    E(w_i) = -\log_2 p(w_i)
    \]
where \( p(w_i) \) is the word's probability in the corpus. 
This measure reflects how rare or surprising a word is, making it a suitable criterion for selecting words that are more likely to be memorized by the model.
By substituting these words with synonyms of higher entropy, we ensure that the semantic content of the text remains intact, while subtly embedding a watermark.

The watermarking process consists of these steps:
\begin{enumerate}
    \item Preprocess Text - The original text is divided into sentences.
    \item Select High-Entropy Words - For each sentence, the top-$K$ words with the highest entropy scores are chosen.

    \item Find High-Entropy Synonyms - Synonyms are retrieved for each of the high-entropy words selected in the previous step, using a specified synonym retrieval method (e.g., BERT, Sentence-BERT (SBERT), or GPT-4o).
    \item Replace Words with Synonyms - Each high-entropy word is replaced by a synonym with a higher entropy score while ensuring that the watermark remains consistent with the original context. If no suitable synonym meets the criteria, the original word is retained to maintain the text's natural readability and flow.
\end{enumerate}

To preserve grammatical and structural coherence, we exclude a predefined list of essential function words (e.g., "a," "an," "the") from modification, while safeguarding the semantic integrity of the text by avoiding alterations to named entities, detected using spaCy~\cite{honnibal2017spacy}, ensuring that key information and meaning remain intact.
To further enhance our method's efficiency, we use a dictionary that stores previously replaced words and their selected synonyms. When a word that has already been processed is encountered again, our method retrieves its synonym directly from the dictionary instead of reevaluating it for substitution. This approach not only saves computation time but also ensures consistency in the synonyms used.

In Figure~\ref{fig:watermark_embedding}, we present an example of the watermark embedding process, illustrating how high-entropy words are replaced with synonyms.
The full embedding algorithm is outlined in Algorithm~\ref{alg:watermark-embedding}.

\begin{algorithm}
\caption{Watermark Embedding Algorithm}
\label{alg:watermark-embedding}
\KwIn{Original text $T$, number of words $K$}
\KwOut{Watermarked text $T_W$}

Split $T$ into sentences and store in $T_W$;\\
\ForEach{sentence $s$ in $T_W$}{
    $H \gets \text{Top-}K \text{ high-entropy words in } s$;\\
    \ForEach{word $h$ in $H$}{
        Find a synonym $h'$ with a higher entropy;\\
        \If{$h'$ exists}{
            Replace $h$ with $h'$;
        }
    }
}
\Return{$T_W$}
\end{algorithm}

\subsubsection*{Synonym Identification Methods}

In this work, we explored several methods for identifying synonyms within text to improve the watermark embedding process. The primary approaches evaluated include WordNet~\cite{WordNet}, BERT~\cite{devlin2019bertpretrainingdeepbidirectional}, and SBERT~\cite{Reimers2019SentenceBERTSE}. 
A detailed runtime comparison of these methods, including their computational overhead, is provided in Appendix~\ref{app:synonym_runtime}.
WordNet functions as a traditional lexical database, offering synonyms without considering context. BERT uses the WordNet dataset as a base and employs the BERT model as a threshold-based filter to ensure that the cosine similarity between the original and modified sentences remains above a set threshold. SBERT further enhances this process by utilizing sentence embeddings from pretrained transformers, allowing it to capture deeper contextual relationships between words and their synonyms.

Additionally, we explored two BERT-based lexical substitution methods: lexical substitution concatenation~\cite{qiang2020lexicalsimplificationpretrainedencoders}, which masks the target word within the sentence and uses BERT to predict the masked token, generating candidate substitutions; and lexical substitution dropout~\cite{zhou-etal-2019-bert}, which applies dropout to the target word's embedding, partially masking the word and validating substitutions based on their effect on the global contextual representation of the sentence. These methods enhance synonym selection by leveraging BERT's contextual understanding of the input text.
For the implementation of these methods, we utilized publicly available code from GitHub~\footnote{https://github.com/jvladika/Lexical-Substitution/tree/main} that uses RoBERTa~\cite{liu2019roberta} as the base model.

For the most accurate synonym generation where the semantic integrity of the sentence is also preserved, we found that GPT-4o~\cite{openai2024gpt4technicalreport} delivered the best results. However, using GPT-4o requires sending sensitive data over the Internet, which raises privacy concerns; therefore, we recommend using a similarly strong language model locally to avoid exposing sensitive data to third parties. More details about the aspect of semantic preservation are provided in Section~\ref{sec:Semantic Preservation}.

\subsection{Watermark Detection}
In the \textbf{watermark detection} phase, our method determines whether the watermarked text was used to train the model by performing an MIA.
Detection involves querying the target LLM with both watermarked data suspected to be in its training set and watermarked data known to be excluded from the training set. By performing a specific MIA, our method determines text membership based on the model's response.
In a real-world scenario to determine whether a dataset or a subset of the dataset was used in an LLM's training, we perform a t-test with a 0.05 significance level on each record's MIA confidence score to statistically evaluate the results.

To determine the best MIA for detecting our watermarked data, we compared our method's performance when the following MIAs were employed: \textit{PPL}~\cite{loss_attack}, \textit{Zlib}~\cite{carlini2021extracting}, \textit{Min-K\%}~\cite{shi2024detectingpretrainingdatalarge} and \textit{Min-K\%++}~\cite{zhang2024minkimprovedbaselinedetecting}.
Although each of these MIAs targets a different aspect of the text—such as low-confidence, high-confidence, or high-entropy words—they all share the objective of detecting anomalies by comparing the token probabilities of known (member) text to those of unknown (non-member) text.

\section{Experimental Setup}
\label{sec:Evaluation}

In this section, we describe the experimental setup used to evaluate \methodname. We conducted experiments on multiple datasets and pretrained LLMs. 

\begin{algorithm}
\caption{Experimental Procedure for Evaluating Watermarking}
\label{alg:Experiment_Setup}
\KwIn{Dataset $D$, Watermarking function $W$, Base LLM $M_{base}$, MIA method $MIA$}
\KwOut{Detection Results}
\SetKwFunction{FMain}{Main}
\SetKwProg{Fn}{Function}{:}{}
\Fn{\FMain}{
    $D_{split} \gets \text{Split}(D)$ \tcp*{Split long records into suitable sizes}
    $D_{member}, D_{non-member} \gets \text{Partition}(D_{split})$ 
    $D_{member} \gets W(D_{member})$ 
    $D_{non-member} \gets W(D_{non-member})$
    $M \gets \text{Fine-Tune}(M_{base}, D_{member})$
    $detection\_results \gets MIA(M, D_{member}, D_{non-member})$\\
    \Return{$detection\_results$}
}
\end{algorithm}

Algorithm~\ref{alg:Experiment_Setup} outlines the steps performed in our experiments to assess watermarking techniques using MIAs for evaluation. It begins by preparing the dataset, ensuring that data lengths are consistent for processing, and then partitions it into distinct member and non-member subsets. Watermarking is subsequently applied to both subsets to assess the resilience of the method under realistic conditions. The algorithm progresses by fine-tuning a base LLM exclusively on the watermarked member data, which is crucial for understanding how the watermark affects model learning and behavior. Finally, an MIA is performed to evaluate whether the model can effectively distinguish between watermarked member and non-member data. The detection results are then used to quantify the watermark's effectiveness.  
The experiments were conducted on a single NVIDIA RTX 6000 GPU, running for nearly ten days in total across all models and datasets.

\subsubsection*{Datasets}
We used six datasets, each comprising distinct types of textual data, commonly used for evaluating MIAs on pretrained LLMs: the \textbf{BookMIA} ~\cite{shi2024detectingpretrainingdatalarge} and five subsets drawn from \textbf{The Pile} ~\cite{gao2020pile800gbdatasetdiverse},
ensuring diverse text types for comprehensive evaluation.

The \textit{BookMIA} dataset consists of 10,000 book snippets, divided into two categories: \textbf{member} and \textbf{non-member} records. Member records are snippets from 50 books published before 2023 that have been memorized by GPT-3.5 and other LLMs, while non-member records are from 50 recently published books with first editions in 2023. 
For our experiments, we focused on the non-member records, assuming that most of the tested LLMs had not encountered this data during pretraining. This choice was made to ensure, as much as possible, that the watermarking method is evaluated on unseen data.

For the \textit{The Pile} dataset, we used the validation set, which was excluded from the pretraining data of the Pythia models~\cite{biderman2023pythiasuiteanalyzinglarge}. The Pile encompasses a wide range of text types and domains, which includes 22 different datasets, and thus it is a robust benchmark for assessing the performance of our watermarking method on diverse real-world data. To ensure a comprehensive evaluation, we selected five datasets from the \textit{The Pile}, each representing a different domain or subject matter. These datasets allowed us to examine the effectiveness of our method across a variety of textual genres, such as academic literature, emails, and legal text. An overview of the datasets is provided in Table~\ref{tab:pile_datasets}, which highlights their diversity.

\begin{table}[h]
\caption{Overview of the Pile datasets used in the Evaluation}
\centering
\footnotesize
\begin{tabular}{c c c}
\hline
\textbf{Dataset}        & \textbf{Content Type} & \textbf{Number of Records} \\ \hline
PubMed Abstracts        & Medical Texts         & 29,871                 \\
Enron Emails            & Emails                & 947                  \\
FreeLaw                 & Legal Documents       & 5,094                  \\
Wikipedia (en)          & Encyclopedic Text     & 17,478                 \\
USPTO Backgrounds       & Patents               & 11,387                 \\
\hline
\end{tabular}

\label{tab:pile_datasets}
\end{table}

\subsubsection*{Models}

We evaluated the performance of \methodname ~on seven pretrained LLMs.

The larger models—\textbf{LLaMA-1 7B}~\cite{touvron2023llamaopenefficientfoundation}, \textbf{LLaMA-3 8B}~\cite{dubey2024llama3herdmodels}, \textbf{Mistral-7B}~\cite{jiang2023mistral7b}, and \textbf{Pythia-6.9B}~\cite{biderman2023pythiasuiteanalyzinglarge}—were fine-tuned on watermarked data using the QLoRA technique~\cite{dettmers2023qloraefficientfinetuningquantized}, which enables efficient training by quantizing model weights to 4-bit precision. Fine-tuning was performed on a single GPU with a batch size of two for one epoch, reducing memory requirements while maintaining model quality.

Additionally, we evaluated continued pretraining on three smaller models: \textbf{Pythia-160M}, \textbf{Pythia-410M}, and \textbf{Pythia-1B}~\cite{biderman2023pythiasuiteanalyzinglarge}. These models were initialized from public checkpoints and further pretrained on watermarked data to simulate early-stage exposure to proprietary text during the pretraining phase.

\begin{table*}[htbp]
\centering
\caption{Comparison of watermarking and non-watermarking methods on various datasets and models based on the AUROC and TPR@FPR=5\% metrics. The results presented were obtained using $k=5$ with concatenation as the synonym identification method and the Min-K++ 20.0\% MIA. Bold values indicate the best performance for each dataset-model pair.}
\label{tab:multi-model-results-pile}
\resizebox{\textwidth}{!}{
\begin{tabular}{cc|ccc|ccc|ccc|ccc}

\toprule
\textbf{Dataset} & \textbf{Metric} & \multicolumn{3}{c}{\textbf{Pythia-6.9B}} & \multicolumn{3}{c}{\textbf{LLaMA-1 7B}} & \multicolumn{3}{c}{\textbf{LLaMA-3 8B}} & \multicolumn{3}{c}{\textbf{Mistral-7B}} \\ \cmidrule(lr){3-5} \cmidrule(lr){6-8} \cmidrule(lr){9-11} \cmidrule(lr){12-14} 
 & & \textbf{None} & \textbf{\methodname} &  & \textbf{None} & \textbf{\methodname} &  & \textbf{None} & \textbf{\methodname} &  & \textbf{None} & \textbf{\methodname} &  \\ \midrule
\multirow{2}{*}{BookMIA} & AUROC & 69.1 & \textbf{94.8} & & 73.2 & \textbf{95.9} & & 79.0 & \textbf{96.9} & & 84.7 & \textbf{96.7} & \\ 
 & TPR@FPR=5\% & 13.5 & \textbf{79.1} & & 18.3 & \textbf{84.3} & & 24.3 & \textbf{84.4} & & 30.2 & \textbf{90.9} & \\ 
\midrule
\multirow{2}{*}{Enron Emails} & AUROC & 65.6 & \textbf{72.3} & & 65.6 & \textbf{69.8} & & 71.3 & \textbf{75.3} & & 78.1 & \textbf{81.6} & \\ 
 & TPR@FPR=5\% & 11.0 & \textbf{23.8} & & 11.0 & \textbf{19.4} & & 12.4 & \textbf{21.3} & & 27.7 & \textbf{31.2} & \\ 
\midrule
\multirow{2}{*}{PubMed Abstracts} & AUROC & 68.7 & \textbf{76.0} & & 72.2 & \textbf{80.7} & & 78.4 & \textbf{83.3} & & 83.8 & \textbf{88.7} & \\ 
 & TPR@FPR=5\% & 17.9 & \textbf{25.0} & & 23.6 & \textbf{35.0} & & 35.4 & \textbf{41.5} & & 48.4 & \textbf{58.4} & \\ 
\midrule
\multirow{2}{*}{Wikipedia (en)} & AUROC & 65.5 & \textbf{74.5} & & 63.1 & \textbf{73.0} & & 70.8 & \textbf{78.9} & & 77.2 & \textbf{84.6} & \\ 
 & TPR@FPR=5\% & 10.2 & \textbf{16.6} & & 12.4 & \textbf{19.7} & & 14.2 & \textbf{22.8} & & 18.1 & \textbf{31.7} & \\ 
\midrule
\multirow{2}{*}{Pile-FreeLaw} & AUROC & 67.7 & \textbf{83.3} & & 57.2 & \textbf{61.6} & & 70.9 & \textbf{87.0} & & 80.1 & \textbf{92.0} & \\ 
 & TPR@FPR=5\% & 10.0 & \textbf{37.0} & & 9.8 & \textbf{23.7} & & 11.8 & \textbf{42.6} & & 18.5 & \textbf{67.1} & \\ 
\midrule
\multirow{2}{*}{USPTO Backgrounds} & AUROC & 63.4 & \textbf{76.1} & & 65.0 & \textbf{78.5} & & 72.4 & \textbf{82.5} & & 82.0 & \textbf{89.8} & \\ 
 & TPR@FPR=5\% & 9.2 & \textbf{22.7} & & 14.5 & \textbf{28.3} & & 21.1 & \textbf{35.2} & & 39.6 & \textbf{60.6} & \\ 

\bottomrule
\end{tabular}
}
\end{table*}


\subsubsection*{Evaluation Metrics}
We evaluated \methodname's performance using two types of metrics: accuracy-related metrics - to assess the effectiveness of watermark detection; and semantic evaluation metrics - to ensure that the original meaning of the text is preserved during watermarking.


\textbf{Accuracy Metrics}: These metrics are used to evaluate how effectively the watermarking method can distinguish between watermarked and non-watermarked data.

\begin{itemize}
    \item \textbf{Area Under the Receiver Operating Characteristic Curve (AUROC)}: The AUROC is a widely used metric for binary classification tasks. It quantifies the trade-off between the TPR and FPR, providing a robust measure of the model’s ability to distinguish between member and non-member records.  
    \item \textbf{True Positive Rate at a fixed False Positive Rate (TPR@FPR)}: This metric is commonly used in classification tasks to measure how effectively positive samples (i.e., watermarked data) are detected, given a fixed rate of false positives. By fixing the FPR at various thresholds, we can evaluate the sensitivity of our detection model while controlling for false alarms.

\end{itemize}

\textbf{Semantic Evaluation Metrics}: These metrics are designed to measure how well the semantic meaning of the text is preserved after the synonym substitution watermarking process has been performed. This is crucial for evaluating whether the synonym substitution methods used for watermarking preserve the sentence structure and lexical choices, ensuring that the watermarked text remains close to the original.

\begin{itemize}
    \item \textbf{Cosine Similarity}: We use both the SBERT model~\cite{Reimers2019SentenceBERTSE} and OpenAI's \textit{text-embedding-3-large}~\footnote{https://platform.openai.com/docs/guides/embeddings} model. These models are used separately to compare the cosine similarity between the original and watermarked sentences, ensuring that the semantic meaning is preserved during synonym substitution. SBERT captures deeper contextual relationships, while \textit{text-embedding-3-large} provides a broader and scalable evaluation, optimized for semantic tasks. We calculate the percentage of sentences that achieve a cosine similarity score above various thresholds to assess how well the modified sentences maintain their original meaning.
    \item \textbf{Bilingual Evaluation Understudy (BLEU) Score}: The BLEU score~\cite{papineni2002bleu} is a well-known metric for evaluating the similarity between a modified text and a reference text (original). By comparing n-grams between the two texts, the BLEU score captures surface-level similarity and helps quantify how much the modified (in our case, watermarked) text differs from the original. 
\end{itemize}


\begin{figure}[ht]
    \centering
    \includegraphics[width=0.9\columnwidth]{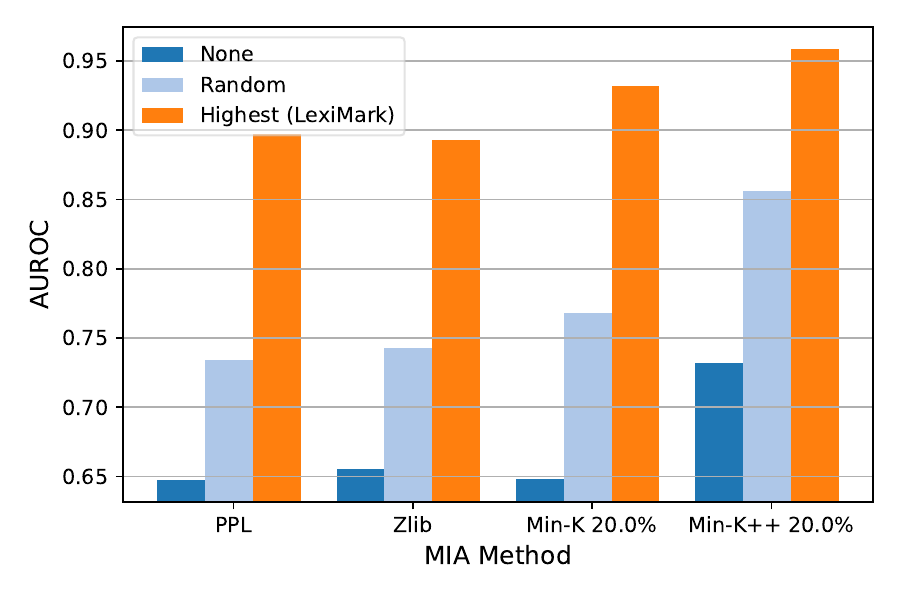}
    \caption{AUROC scores obtained using different watermarking techniques on the BookMIA dataset with the LLaMA-1 7B model. Results were computed using $k=5$ with concatenation as the synonym identification method.}

    \label{fig:auc_bar_graph}
\end{figure}

\section{Results}
\label{sec:results}


In this section, we present the results of the experiments conducted to evaluate \methodname.
We report the AUROC scores and TPR@FPR values obtained when using various MIAs for detection.
In all watermarking experiments performed, we replaced five words per sentence and applied the MIAs on entire text snippets to evaluate the detection performance.
An evaluation examining the use of different \(k\) values is presented in Appendix~\ref{app:k_values}.
Our experiments employ lexical substitution concatenation~\cite{qiang2020lexicalsimplificationpretrainedencoders} with a threshold of five as the synonym substitution method, chosen for its effective balance between watermarking efficiency and semantic preservation. Further details on various synonym methods and their impact on semantic preservation are discussed in Section~\ref{sec:Semantic Preservation}.




\subsection{Fine-Tuning Results (QLoRA Setting)}

Table~\ref{tab:multi-model-results-pile} compares the results of \methodname\ against a baseline approach in which no watermarking is applied. 
The evaluation spans various datasets, and LLMs (Pythia-6.9B, LLaMA-1 7B, LLaMA-3 8B, and Mistral-7B). The reported results correspond to the detection performance when employing the \textit{Min-K++ 20.0\%} MIA, measured in terms of AUROC and TPR@FPR=5\%.

Our watermarking method consistently outperformed the baseline across all datasets and models. For instance, on the BookMIA dataset, our method improved the AUROC from 69.1 to 94.8 with Pythia-6.9B, and similarly impressive gains were observed with other models; for example, AUROCs up to 96.9\% were achieved with LLaMA-3 8B. Similarly, on the Pile-FreeLaw dataset, the TPR@FPR=5\% increased from 10.0 to 37.0 with Pythia-6.9B.
Such improvements are also seen on the other datasets. Notably, on Pile-FreeLaw, our method increased the AUROC from 67.7\% to 83.3\% with Pythia-6.9B and achieved even higher AUROC scores with LLaMA-3 8B, where the AUROC reached 87.0.
On the USPTO Backgrounds dataset, the AUROC increased from 63.4\% to 76.1\% with Pythia-6.9B, and the TPR@FPR=5\% also improved, going from 9.2 to 22.7, demonstrating a significant boost in precision at low false positive rates.

To validate our strategy of replacing high-entropy words with their higher-entropy synonyms and to assess the impact of different MIA methods, we compared \methodname\ against a baseline that randomly replaces words with randomly chosen synonyms.
Figure~\ref{fig:auc_bar_graph} displays the AUROC scores obtained by the different techniques on the BookMIA dataset evaluated with the \textit{Min-K++ 20.0\%} MIA, using the LLaMA-1 7B model. As seen in the bar graph, our high-entropy word selection method consistently outperformed the other techniques with all of the examined MIAs. Without watermarking (None), the MIAs achieved AUROC scores between 63.8\% and 73.1\%, indicating limited ability to detect membership.
Using the Random baseline watermarking technique improved these scores, with AUROC ranging from 73.4\% to 85.6\%.
In contrast, using our high-entropy word replacement watermarking technique, the AUROC scores were consistently above 90\% and when using the \textit{Min-K++ 20\%} MIA as the detection tool it, scores approached nearly 100\%.
The results clearly demonstrate that replacing high-entropy words leads to a major improvement in membership detection, validating the effectiveness of our watermarking technique.

In conclusion, the consistent performance improvements across all examined LLMs, along with substantial gains in the AUROC and TPR@FPR=5\% metrics, highlight the effectiveness, versatility, and robustness of our watermarking technique across diverse datasets, particularly challenging ones like BookMIA. Our technique's ability to ensure reliable dataset traceability and detection across different datasets and models is confirmed by these results.

\subsection{Continued Pretraining Results}

    \begin{table*}[htbp]
    \centering
    \caption{Comparison of watermarking and non-watermarking methods on various datasets and models based on the AUROC and TPR@FPR=5\% metrics. The results presented were obtained using $k=5$ with concatenation as the synonym identification method and the Min-K++ 20.0\% MIA. Bold values indicate the best performance for each dataset-model pair.}
    \label{tab:pythia_pretrained_results}
    \resizebox{\textwidth}{!}{%
    \begin{tabular}{cc|ccc|ccc|ccc}
    \toprule
    \textbf{Dataset} & \textbf{Metric} & \multicolumn{3}{c}{\textbf{Pythia-160M}} & \multicolumn{3}{c}{\textbf{Pythia-410M}} & \multicolumn{3}{c}{\textbf{Pythia-1B}} \\ \cmidrule(lr){3-5} \cmidrule(lr){6-8} \cmidrule(lr){9-11} 
 & & \textbf{None} & \textbf{\methodname} &  & \textbf{None} & \textbf{\methodname} &  & \textbf{None} & \textbf{\methodname} &  \\ \midrule
\multirow{2}{*}{BookMIA} & AUROC & 77.5 & \textbf{95.0} & & 87.3 & \textbf{97.0} & & 88.1 & \textbf{96.2} & \\ 
 & TPR@FPR=5\% & 18.0 & \textbf{89.5} & & 25.0 & \textbf{95.9} & & 24.5 & \textbf{95.0} & \\ 
\midrule
\multirow{2}{*}{Enron Emails} & AUROC & 79.1 & \textbf{85.2} & & 84.6 & \textbf{87.6} & & 85.8 & \textbf{89.0} & \\ 
 & TPR@FPR=5\% & 26.8 & \textbf{51.3} & & 31.0 & \textbf{59.2} & & 48.0 & \textbf{68.4} & \\ 
\midrule
\multirow{2}{*}{PubMed Abstracts} & AUROC & 69.9 & \textbf{77.9} & & 86.5 & \textbf{89.0} & & 93.8 & \textbf{96.5} & \\ 
 & TPR@FPR=5\% & 17.9 & \textbf{26.8} & & 52.9 & \textbf{60.2} & & 82.7 & \textbf{89.8} & \\ 
\midrule
\multirow{2}{*}{Wikipedia (en)} & AUROC & 68.4 & \textbf{74.5} & & 76.8 & \textbf{84.5} & & 80.2 & \textbf{87.9} & \\ 
 & TPR@FPR=5\% & 10.0 & \textbf{17.0} & & 18.1 & \textbf{37.9} & & 33.4 & \textbf{57.1} & \\ 
\midrule
\multirow{2}{*}{PILE-FreeLaw} & AUROC & 67.2 & \textbf{79.8} & & 73.9 & \textbf{87.1} & & 78.1 & \textbf{91.4} & \\ 
 & TPR@FPR=5\% & 13.5 & \textbf{34.5} & & 18.8 & \textbf{46.9} & & 23.4 & \textbf{64.3} & \\ 
\midrule
\multirow{2}{*}{USPTO Backgrounds} & AUROC & 69.5 & \textbf{79.4} & & 80.5 & \textbf{89.8} & & 83.5 & \textbf{92.0} & \\ 
 & TPR@FPR=5\% & 17.4 & \textbf{29.5} & & 41.5 & \textbf{61.2} & & 54.1 & \textbf{73.2} & \\ 

\bottomrule
\end{tabular}
}
\end{table*}

To further validate the watermark’s learnability during early training stages, we evaluated continued pretraining on smaller models: Pythia-160M, Pythia-410M, and Pythia-1B. Table~\ref{tab:pythia_pretrained_results} presents AUROC and TPR@FPR=5\% results on multiple datasets using the \textit{Min-K++ 20.0\%} MIA.

Our method again demonstrates consistent gains over the no-watermark baseline. For example, on PILE-FreeLaw, AUROC improves from 73.9\% to 87.1\% with Pythia-410M. On BookMIA, TPR@FPR=5\% increases from 18.0\% to 89.5\% with Pythia-160M. The largest model, Pythia-1B, achieves up to 96.5\% AUROC.
These results confirm that \methodname~is highly learnable and effective even when embedded early in the pretraining pipeline, reinforcing its applicability for both fine-tuned and pretrained LLM scenarios.




\section{Semantic Preservation}
\label{sec:Semantic Preservation}

One of the most critical aspects of watermarking textual data used to train LLMs is ensuring that the watermarks preserve the meaning of the original text~\cite{fang2023cosywa,kirchenbauer2023watermark}. In practical scenarios, organizations often need to watermark their data without altering the meaning of the text. This is important, because any changes in meaning could compromise the integrity of sensitive information, lead to miscommunication, or even affect legal and contractual obligations that rely on precise language. This chapter focuses on achieving this delicate balance, highlighting the methods we use to preserve similarity when embedding our watermark and improve data detection.

Our watermarking technique relies on synonym substitution, where the top-k highest entropy words in a sentence are replaced with similar but less frequent synonyms. The challenge lies in ensuring that the replacements are semantically close enough to the original words such that the text remains coherent and the meaning is unchanged. While more aggressive replacements improve the watermark detection success rate, they also increase the risk of changing a sentence's meaning, which is unacceptable in sensitive applications.

\subsection{Semantic Evaluation}
In our semantic evaluation, we examined how well different methods, including BERT and SBERT, when used by \methodname ~to select synonyms, preserve the meaning of watermarked text with various cosine similarity thresholds. As the threshold increases from 0.8 to 0.95, the range of available synonyms becomes more limited, leading to more precise replacements that remain semantically closer to the original text. This improves semantic preservation, as shown in Table~\ref{tab:bert_sbert_threshold}. For instance, BERT’s cosine similarity increased from 88.33\% at a threshold of 0.8 to 99.9\% at 0.95; SBERT also showed a dramatic rise, reaching 99.49\% cosine similarity at the highest threshold.

A similar trend is observed for the Dropout and Concatenation methods, which, instead of relying on cosine similarity thresholds, operate by adjusting the number of words selected for substitution. These methods return a list of candidate synonym words ranked by their contextual relevance, whereas our method selects the top-k candidates.
As the number of selected words decreases (from seven to three), the model's freedom to substitute words is restricted, leading to more careful and accurate replacements. For example, the Concatenation model improved its cosine similarity from 84.01\% when selecting the top-7 words to 93.68\% when selecting only the top-3 words, as shown in Table~\ref{tab:bert_sbert_threshold}, underscoring how the selection of fewer words yields better semantic fidelity.

\subsection{Trade-offs Between AUROC and Semantic Preservation}


Our experiments reveal a trade-off between semantic preservation and watermark detection. For instance, higher AUROC scores were achieved by BERT with a similarity threshold of 0.8 than achieved with a 0.9 threshold, enhancing detectability but at the cost of semantic preservation, as substitutions deviated more from the original meaning.

For example, consider the following sentence:

\begin{quote}
\large
\textit{"The board \textbf{discussed} the potential risks associated with the merger."}
\end{quote}
If we replace \textit{"discussed"} with \textit{"debated"} (cosine similarity = 0.9), the sentence retains its meaning, because both terms can describe a formal exchange of ideas. However, if we replace \textit{"discussed"} with \textit{"argued"} (cosine similarity = 0.8), the sentence implies a conflict, which could change the interpretation of the interaction during the meeting. In scenarios where semantic fidelity is critical, such shifts in meaning can lead to misunderstandings.

\begin{table}[htbp]
\centering
\caption{Evaluation of the trade-off between the AUROC, cosine similarity (CosSim), and BLEU score on the BookMIA dataset with the Min-K++ 20.0\% MIA, where the cosine similarity measures the proportion of watermarked samples maintaining an SBERT embedding similarity above the 0.8 threshold.}
\label{tab:bert_sbert_threshold}
\resizebox{0.9\columnwidth}{!}{
\begin{tabular}{ccccc}
\toprule
\textbf{Method} & \textbf{Threshold} & \textbf{AUROC} & \textbf{CosSim} & \textbf{BLEU} \\
\midrule
\multirow{4}{*}{BERT} & 0.8 & \textbf{94.00} & 88.33\% & 0.60 \\ 
                       & 0.85 & 93.80 & 95.02\% & 0.65 \\ 
                       & 0.9 & 92.30 & 99.03\% & 0.73 \\ 
                       & 0.95 & 88.70 & \textbf{99.90\%} & \textbf{0.84} \\ 
\midrule
\multirow{4}{*}{SBERT} & 0.8 & 91.50 & 59.19\% & 0.54 \\ 
                       & 0.85 & 93.70 & 72.85\% & 0.56 \\ 
                       & 0.9 & \textbf{94.50} & 89.46\% & 0.60 \\ 
                       & 0.95 & 94.10 & \textbf{99.49\%} & \textbf{0.69} \\ 
\midrule
\multirow{3}{*}{Dropout} & 7 & 96.50 & 47.46\% & 0.48 \\ 
                       & 5 & \textbf{96.80} & 59.17\% & 0.51 \\ 
                       & 3 & 96.50 & \textbf{77.14\%} & \textbf{0.56} \\ 
\midrule
\multirow{3}{*}{Concatenation} & 7 & \textbf{96.20} & 84.01\% & 0.52 \\ 
                       & 5 & 95.90 & 88.57\% & 0.53 \\ 
                       & 3 & 95.10 & \textbf{93.68\%} & \textbf{0.57} \\ \bottomrule
\end{tabular}
}
\end{table}

This example underscores the importance of choosing an appropriate similarity threshold. As shown in Table~\ref{tab:bert_sbert_threshold}, although lower thresholds (e.g., 0.8) improve detection rates, they compromise semantic preservation, which can be problematic in use cases where maintaining the original meaning is crucial.

\subsection{Optimizing Semantic Preservation}
In our effort to balance the accuracy of the watermark detection with semantic preservation, it became clear that using similarity thresholds of 0.8 or 0.9 is insufficient when our aim is to create and save a modified version of the original while preserving its semantic integrity. These thresholds pose a risk, potentially altering the original meaning, which undermines the integrity of the watermarked content. To address this problem, we use higher similarity thresholds (e.g., 0.95). We evaluated BERT and Sentence-BERT (SBERT) models, using a higher cosine similarity threshold of 0.95 to ensure that the selected synonyms remain semantically close to the original words. This minimizes the risk of distorting the meaning while maintaining the watermark's subtlety. We further explored GPT-4o, a more advanced language model, to select higher-entropy synonyms, offering a superior approach to improving the watermark's subtlety and effectiveness while preserving readability. 
Although GPT-4o was chosen for this task due to its advanced capabilities, it relies on a remote API and does not ensure data privacy in sensitive applications; however, our method is adaptable and can be applied locally with other LLMs to address privacy concerns.

\begin{figure}[ht]
    \centering
    \includegraphics[width=0.9\columnwidth]{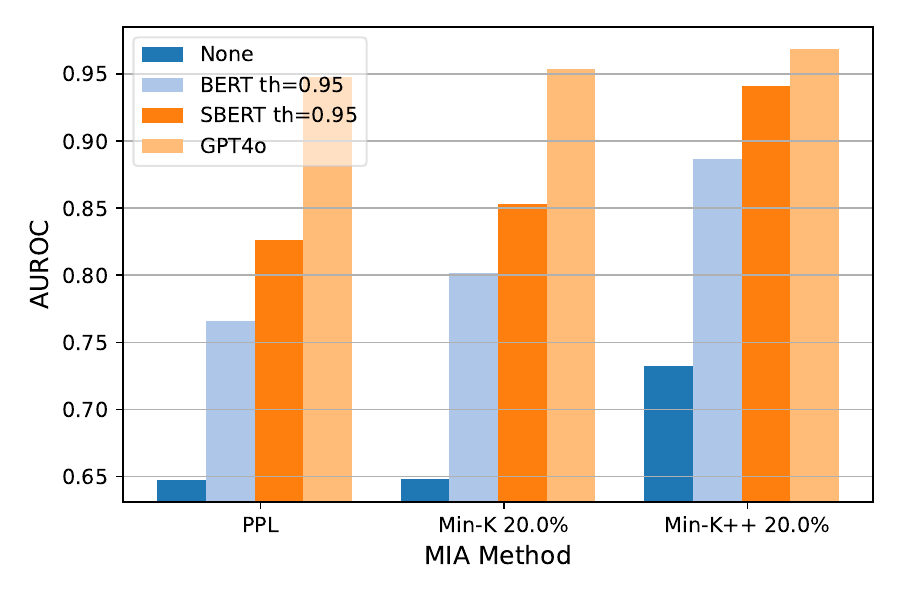}
    \caption{AUROC scores comparing various synonym identification methods for watermark detection on the BookMIA dataset, highlighting the method with the highest semantic preservation.}

    \label{fig:bookMIA_gpt}
\end{figure}

The results presented in Figure~\ref{fig:bookMIA_gpt} demonstrate our method's ability to achieve strong watermark detection results, even with this restrictive threshold.
The AUROC score for GPT-4o reached almost 95\% for all attacks, with very strong performance using the \textit{Min-K++ 20\%} method, where it achieved a detection success rate of 97\%. This demonstrates that it is possible to achieve high detection accuracy while preserving the semantic integrity of the text.

\begin{figure}[ht]
    \centering
    \includegraphics[width=0.9\columnwidth]{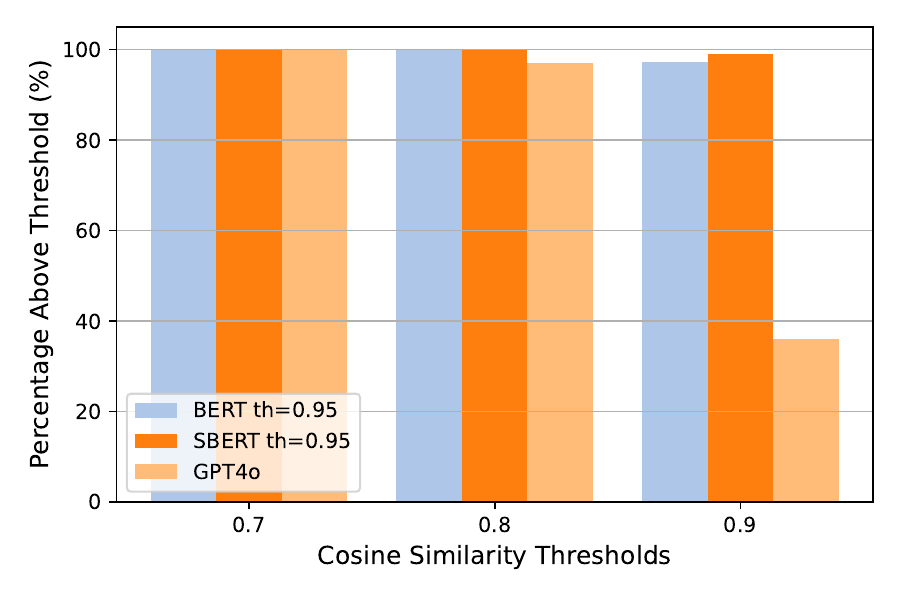}
    \caption{Semantic similarity evaluation on the BookMIA dataset using the GPT embedding model "text-embedding-3-large," showing the proportion of watermarked samples with cosine similarity above various thresholds.}
    \label{fig:semantic_similarty_gpt}
\end{figure}

To measure semantic preservation in this case, we utilized OpenAI's text-embedding-3-large model, leveraging its advanced capabilities as described in Section~\ref{sec:Evaluation}. We explored cosine similarity thresholds of [0.7, 0.8, 0.9], with the results clearly illustrating the effect of each threshold on maintaining the semantic integrity of the watermarked text, as shown in Figure~\ref{fig:semantic_similarty_gpt}.

As seen in the figure, BERT and SBERT consistently outperformed both GPT-4o methods in preserving the meaning of the text, as indicated by their higher semantic scores. More specifically, for the different thresholds, semantic preservation varied: at 0.7, all models preserved the meaning completely (100\%); at 0.8, BERT and SBERT maintained a score of 100\%, while GPT-4o dropped slightly to a score 97\%; and at 0.9, SBERT and BERT retained high scores of 98\% and 97\% respectively, while GPT-4o performed poorly, falling down to a score of 36\%.

Upon closer examination, it became evident that the lower scores of the GPT-4o method were likely due to the fact that it replaced more words per sentence (on average four to five words were replaced) compared to BERT and SBERT with a similarity threshold (th) of 0.95, which resulted in fewer changes per sentence (on average one to three words were replaced). This suggests that the lower semantic scores for GPT-4o may be attributed to the fact that its watermarked sentences contained fewer words from the original sentence than those produced by BERT and SBERT.

\subsection{Alternative Use Cases for Lower Similarity Thresholds}
\label{sec:similarity_backdoor}
While higher thresholds, such as 0.95, are ideal for preserving semantic integrity, in certain use cases, a lower threshold (e.g., 0.8) offers a unique advantage. Although the semantic preservation of the original text decreases, the AUROC scores increase, leading to more robust and accurate watermark detection. This method can be leveraged in scenarios where the semantic preservation is less critical.

One practical application of using lower thresholds is to create honeypot text files in our dataset with low similarity thresholds. By watermarking non-sensitive texts with a lower threshold (e.g., 0.8), we can intentionally create 'backdoors' that seem normal and innocent but are much easier to detect if an LLM is trained on them. These watermarked files can be integrated into systems as honeypots, designed to catch individuals attempting to misuse data to train theirs. Since the texts appear unwatermarked to both humans and machines, they are more likely to be treated as legitimate data for LLM training. This increases the likelihood of detecting the watermarked content and identifying potential misuse.
As mentioned in Section~\ref{sec:dataset_detection}, we found that the use of as few as six records is sufficient for determining the membership status of an entire dataset, suggesting that using only a small percentage of the data can be effective.
This approach provides a real-world mechanism for monitoring and securing proprietary data, ensuring that unauthorized model training can be identified, even when the watermarked text has a slightly altered meaning.

\section{Robustness}
\label{sec:robustness}
\label{sec:Combined}

In this section, we explore the robustness of our watermarking method and compare it to two existing approaches used for watermarking in LLM training: the \textit{random sequence watermark} and \textit{Unicode watermark}~\cite{wei2024proving} methods.
Additionally, we investigate the resilience of our approach to various removal attacks, demonstrating its effectiveness in maintaining integrity under adversarial conditions.

\subsection{Detectability}
One of the key factors for a watermarking method is its detectability~\cite{liang2024watermarkingtechniqueslargelanguage}. 
\methodname ~is highly resistant to detection, as it only uses lexical substitutions that maintain the sentence structure and preserve the original meaning.

There are several common approaches for watermarking text. One such approach is the \textbf{random sequence watermark} method, which inserts randomly generated sequences into the text, making it easily detectable by human readers or a simple filter function.
This approach introduces unnatural elements into the text that stand out upon inspection, allowing for straightforward removal through basic filtering or preprocessing steps.

Another approach is the \textbf{Unicode watermark} method, where certain characters are replaced with visually identical Unicode characters. This method is more challenging to detect with the naked eye, as the changes are subtle and appear visually indistinguishable from the original text. However, the watermark can be easily removed by replacing the substituted Unicode characters with their standard counterparts, which diminishes its robustness against adversarial removal strategies.
Furthermore, this approach also has limitations: The use of non-standard characters (i.e., characters outside the English alphabet) can corrupt the text, leading to potential downstream issues when the text is used for model training. Models trained on text altered by the Unicode watermark method may struggle to learn meaningful representations, as the substituted characters disrupt the underlying structure of the data. 
One clear sign of such disruption is the increase in perplexity—a measure of how well a model predicts the next token in a sequence. 
When trained on watermarked text, the model's perplexity is often higher compared to when trained on clean data, as the model faces difficulty in accurately predicting sequences due to the altered characters~\cite{carlini2023quantifyingmemorizationneurallanguage,kirchenbauer2023watermark}.

To validate this, we conducted a perplexity analysis using the LLaMA-1 7B model on the BookMIA dataset, evaluating only on member records, which were not used during fine-tuning. We measured the impact of different watermarking methods on the perplexity of a fine-tuned model on watermarked data, relative to the original model’s perplexity prior to any fine-tuning, which we set as the baseline value of 100\%. 
To calculate the perplexity ratio (PR), we used the formula:
\[
\text{PR} = \left( \frac{\text{Perplexity of Original Model}}{\text{Perplexity of Fine-tuned Model}} \right) \times 100
\]
Unlike standard perplexity, where lower values indicate better performance, a higher PR value (closer to 100\%), indicates better preservation of the original model's performance on the examined text.
Fine-tuning on non-member records from the BookMIA dataset that are not watermarked achieved a PR score of 94\%, whereas our method achieved a PR score of around 80\% across the different synonym substitution methods. Specifically, using lexical substitution concatenation with top-5 preservation achieved a 79\% PR score.
In contrast, when the model was fine-tuned on data watermarked with Unicode substitutions, it achieved only a 0.0005\% PR score, indicating a significant decrease in performance.
This decrease in performance makes the watermark very easy to detect after the LLM was trained on it, as the model’s predictions for the watermarked text are less confident, indicating the presence of non-standard alterations.

\begin{table}[ht]
\centering
\footnotesize
\caption{Comparison of baseline watermarking methods in terms of detectability and ease of removal.}
\label{tab:baseline_comparison}
\begin{tabular}{lcc}
\toprule
\textbf{Method} & \textbf{Detectability} & \textbf{Ease of Removal} \\ 
\midrule
Random Seq  & Easy     & Easy    \\ 
Unicode & Easy          & Medium         \\ 
Ours (\methodname)  & \textbf{Hard}     & \textbf{Hard}         \\
\bottomrule
\end{tabular}
\end{table}

The comparison provided in Table~\ref{tab:baseline_comparison} highlights the advantage of our \methodname ~method over existing approaches. While both the \textit{random sequence watermark} and \textit{Unicode watermark} methods are easily detectable and removable, \methodname ~stands out as being highly resistant to detection and considerably harder to remove. This demonstrates \methodname's robustness in embedding watermarks without compromising the text's integrity or introducing detectable artifacts, making it a far more secure and reliable option for watermarking.

\subsection{Combined Watermark Evaluation}
Both the \textit{random sequence watermark} and \textit{Unicode watermark} methods use distinct detection techniques and metrics to assess their robustness. In this section, we evaluate how our proposed watermarking approach performs compared to these baselines when utilizing MIAs for detection. Additionally, we explore the potential advantages of combining our method with these existing techniques. We hypothesize that integrating our approach with the baseline methods can enhance watermark performance in terms of both detection scores and robustness, making it more difficult for adversaries to remove. Even if the simpler watermarks like the \textit{random sequence watermark} and \textit{Unicode watermark} methods are detected and eliminated, our watermark will remain intact, providing an additional layer of security.

\begin{figure}[ht]
    \centering
    \includegraphics[width=0.9\columnwidth]{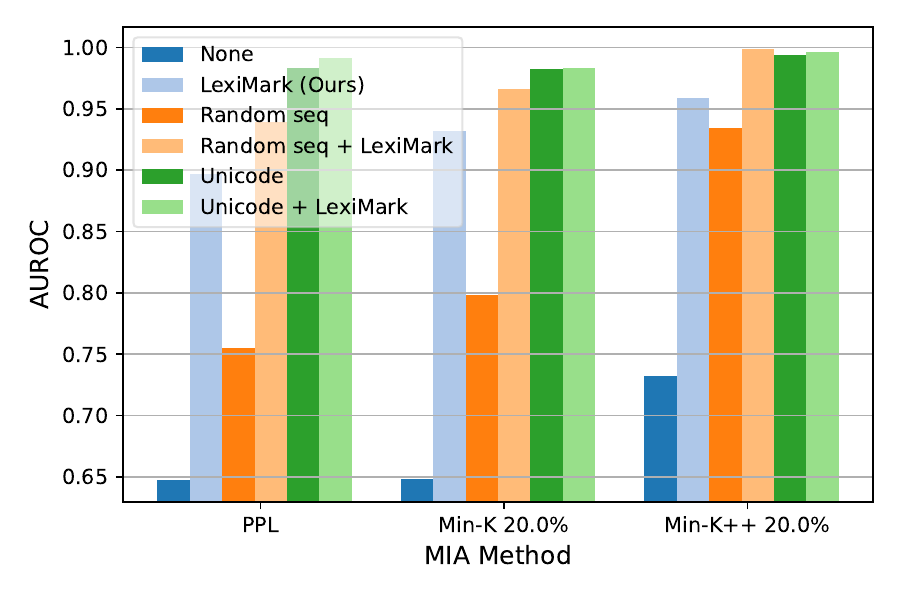}
    \caption{AUROC scores comparing various watermarking methods, focusing on combined approaches, on the BookMIA dataset, using the LLaMA-1 7B model, with k=5 using concatenation as the synonym identification method.}
    \label{fig:bookMIA_combined}
\end{figure}

Figure~\ref{fig:bookMIA_combined} presents the results of the MIA method applied to both the baseline techniques and the combination with \methodname. Our approach outperforms the \textit{random sequence watermark} and is slightly behind the \textit{Unicode watermark} method in standalone comparisons. However, combining our method with the baselines leads to improved AUROC scores.

Combining the \textit{random sequence watermark} method with \methodname ~results in an AUROC improvement ranging from 6.5\% to 18.4\%. Using the \textit{Min-K 20\%} as the detection tool, the AUROC increases from 79.9\% to 96.6\%. 
We can further see evidence supporting our hypothesis with the combination of \methodname ~and the \textit{Unicode watermark} method. While the \textit{Unicode watermark} method already achieves strong results on its own, integrating it with \methodname ~yields a modest 1\% AUROC improvement.

\subsection{Robustness to text modification}
In this section, we evaluate the robustness of \methodname ~against common text modifications, focusing on its resilience to synonym substitution attacks. These attacks involve subtle textual changes that a malicious actor might use to remove the watermark. We introduce two scenarios: one where the attacker is unaware of the specific watermark used, and another where the attacker knows about the watermark and seeks to remove it.

\subsubsection*{Random Synonym Substitution Attack}
In the first scenario, we simulate an attack where the dataset, already embedded with our watermark, is modified by randomly replacing $K$ words in each sentence with their synonyms. This modification simulates an adversary's actions, where, unaware of the specific watermark, they aim to alter the text to reduce the success rate of our watermark detection. The synonym-substituted dataset is then used to train an LLM

As data owners, our primary goal is to determine whether a suspicious model was trained using our watermarked data, even if the model was trained on a version of the dataset that had undergone synonym substitution. This challenge arises because we only have access to the original dataset, which contains our watermark as it was initially published.


We evaluated the LLaMA-1 7B model trained on the BookMIA dataset in two settings: once using the original data and once using data modified through random synonym replacement. 
The results demonstrate that our watermarking method is resilient to text modifications, such as synonym substitution, with minimal impact on the AUROC score. The \textit{PPL} and \textit{Zlib} methods experience the largest decreases in AUROC—5.90\% and 6.00\%, respectively. In contrast, the \textit{ Min-K++ 20.0\%} method exhibits the greatest resilience, with only a 2\% reduction, and the\textit{ Min-K 20.0\%} method follows closely with a 4.40\% drop. 
Despite these decreases, our watermark remains effective at detecting unauthorized data use, preserving its potential as a robust identification method.

\subsubsection*{Targeted High-Entropy Synonym Substitution Attack}
In the second attack scenario, the attacker targets the $K$ highest entropy words for replacement with their low-entropy synonyms in an effort to remove our watermark. 
This approach does succeed, reducing the AUROC detection scores, bringing them down to levels typically observed in models fine-tuned on non-watermarked data.
As outlined in Section~\ref{sec:similarity_backdoor}, we can strategically apply the watermark to only a few samples to minimize its impact on model perplexity while maintaining a high detection rate. 
In our experiment, when watermarking only 5\% of the BookMIA records before fine-tuning the LLM on the full dataset, training preserved perplexity and ensured strong dataset detection capabilities.
Our approach achieved a 90.82\% PR score, whereas the attacker's model achieved only a 76.21\% PR score. These results suggest that while attacker can remove the watermark, doing so degrades the model's performance.

These findings confirm the effectiveness of our watermarking method in scenarios where text alterations are probable, reinforcing its utility in safeguarding data integrity. Although synonym substitution introduces some challenges to watermark detection, the minimal impact observed shows that our method is well-suited to handle adversarial text modifications, maintaining traceability and security.

\subsection{Robustness to Post-Training}

We evaluate the persistence of our watermark after subjecting the model to additional post-training, which typically occurs in multiple phases, as described below.

\subsubsection{Continued Pretraining}

In this experiment, we assess whether our watermark remains detectable after the model undergoes further training on a new dataset. Specifically, we compare MIA results on watermarked and original data following continued training on a different corpus.

We evaluate two models:

\begin{itemize}
    \item LLaMA-3 8B, which is first fine-tuned using QLoRA on the BookMIA dataset (both original and watermarked versions used as suspect records), and then further fine-tuned on the Enron Emails dataset.
    \item Pythia-410M, which undergoes standard pretraining (rather than QLoRA-based fine-tuning) on BookMIA followed by continued pretraining on the Enron Emails dataset.
\end{itemize}

For LLaMA-3 8B, we observe a modest drop in MIA performance when using \methodname, from an AUROC of 96.9\% to 90.6\% with the MIN-K++ 20.0\% MIA method. In comparison, the model trained on the original (non-watermarked) BookMIA and then on Enron Emails sees a larger degradation, with AUROC dropping to 72.6\%.

For Pythia-410M, using \methodname, the AUROC drops from 97.0\% to 86.7\% after continued pretraining. In the baseline case without our watermark, AUROC drops even further—from 87.3\% to 76.2\%.

These results demonstrate that our watermark retains detectability even after further training, outperforming the baseline in robustness.

\subsubsection{Instruction Tuning}
Instruction tuning modifies a model’s behavior to better align with human-provided prompts and objectives, which may influence its ability to retain previously embedded watermark signals. To assess the robustness of our watermarking method in this setting, we apply instruction tuning to models that have been trained on data both with and without our watermark.

We evaluate this scenario using the Pythia-410M model, which first undergoes standard pretraining on the BookMIA dataset, followed by instruction tuning on the TriviaQA dataset~\cite{Joshi2017TriviaQAAL}. After instruction tuning, the AUROC of our method using the MIN-K++ 20.0\% MIA drops slightly from 97.0\% to 93.7\%, indicating that the watermark remains highly detectable. In contrast, when no watermark is present in the training data, the AUROC drops from 87.3\% to 82.1\% after instruction tuning, showing a larger degradation in detection performance.

\section{Dataset Detection}
\label{sec:dataset_detection}

LLM Dataset Inference is a more recent and relevant evaluation approach than single-record detection for identifying whether an entire dataset or portions of it were used in model training~\cite{maini2024llm}. Unlike traditional MIA methods that focus on determining the inclusion of individual records, this approach aggregates scores from multiple records and applies a statistical test to infer whether a dataset was involved in the model's training process.

In our dataset inference evaluation, we aimed to identify the minimum number of member and non-member records required to reliably conduct a statistical t-test, ensuring a p-value of below 0.05. We iterated over group sizes ranging from two to 100 records for both member and non-member sets. For each group size, we randomly sampled records from each set and performed a statistical test on the scores generated by the MIA. This process was repeated 100 times for each group size, and we calculated the average p-value across all iterations.

\begin{figure}[ht]
    \centering
    \includegraphics[width=0.9\columnwidth]{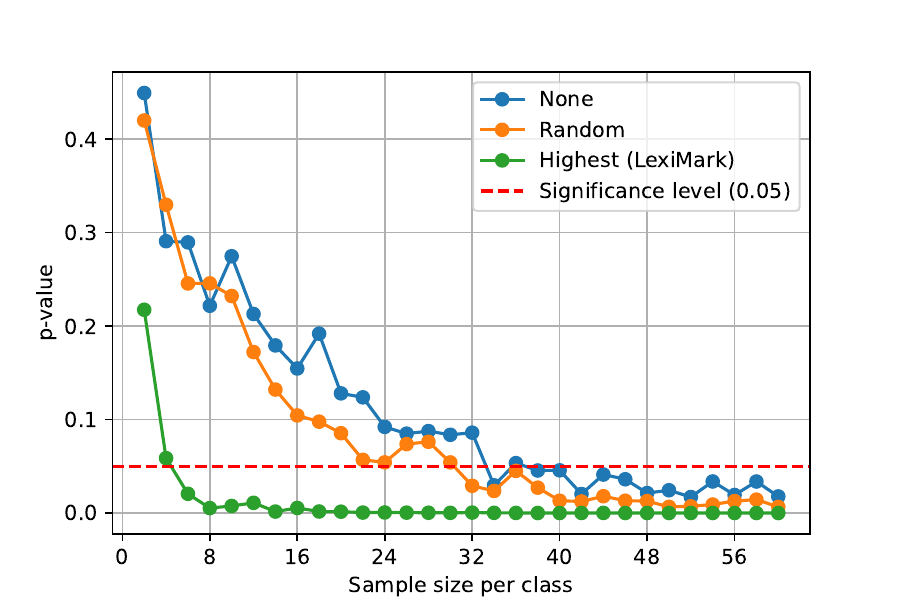}    
    \caption{Average p-value as a function of the group size, comparing member and non-member records using the LLaMA-1 7B model fine-tuned on the BookMIA dataset, with the \textit{Min-K++ 20\%} MIA.}
    \label{fig:bookMIA_p_value}
\end{figure}

Figure~\ref{fig:bookMIA_p_value} presents the average p-value as a function of the number of records sampled from each group. The results are based on the LLaMA-1 7B model fine-tuned on the BookMIA dataset, using the \textit{Min-K++ 20\%} method as the MIA.
The methods use lexical substitution concatenation~\cite{qiang2020lexicalsimplificationpretrainedencoders} as the synonym substitution technique. 

As shown, our method achieves an average p-value below 0.05, with as few as six records per group, indicating statistical significance very close to zero. In contrast, for data without any watermarking, at least 40 records per group are required to reach statistically significant results. This highlights the efficiency of our method in conducting reliable dataset inference with smaller sample sizes.

In real-world scenarios, when a data owner suspects that a model has been trained on their data, they often cannot determine what percentage of the data was used for training. To evaluate this scenario, we present the results of dataset detection when the model was trained on only a portion of the member data. This simulates a common scenario where the data owner possesses non-member data that includes recent or evolving content that has not yet been published or made publicly available.

\begin{figure}[ht]
    \centering
    \includegraphics[width=0.9\columnwidth]{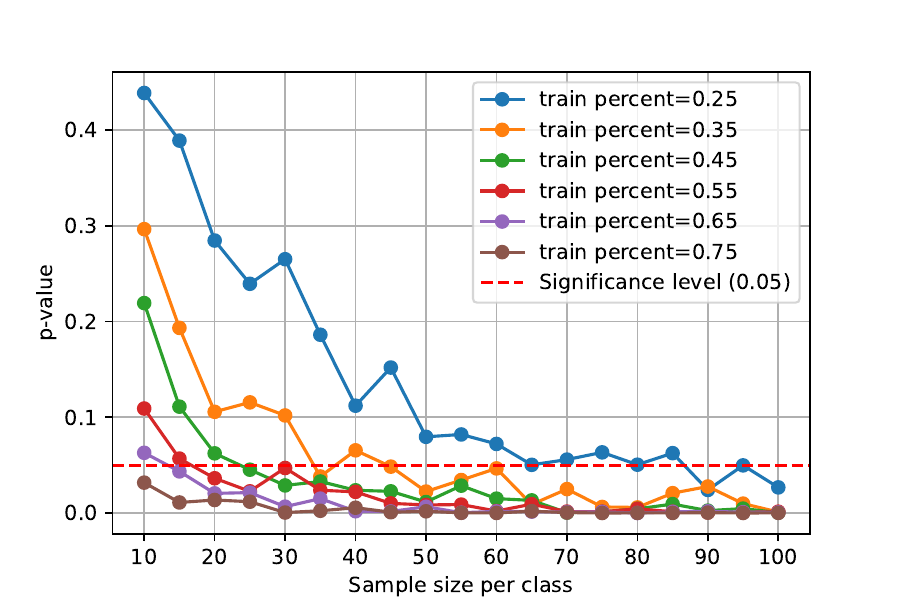}
\caption{Average p-value as a function of group size, where only a subset of the member data is used for training. The results are shown for different percentages of member data on the LLaMA-1 7B model fine-tuned on the BookMIA dataset, using the Min-K++ 20\% MIA.}

\label{fig:bookMIA_p_value_partial}
\end{figure}

Figure~\ref{fig:bookMIA_p_value_partial} presents the results of the LLaMA-1 7B model fine-tuned on the BookMIA, dataset using the \textit{Min-K++ 20\%} method as the MIA, indicating the number of records needed from both the member and non-member groups to achieve statistical significance. Each line in the graph, represented by different colors, indicates the percentage of member records used to train the model. The results are averaged across 100 iterations, with group sizes ranging from 10 to 100 in steps of five.

As observed in the figure, when the model is trained on only 35\% of the member data, sampling 50 member records and 50 non-member records (which are known not to have been used for training) is sufficient to achieve a p-value below 0.05, indicating statistical significance. This demonstrates that even when the model has been trained on only a subset of the member data, it is possible to detect whether the model has been exposed to this subset of data.



\section{Conclusion}
\label{sec:conclusion}

In this paper, we presented \methodname, a novel watermarking technique designed to improve the detection of datasets used to train LLM. \methodname ~uniquely embeds watermarks by substituting high-entropy words with their synonyms, ensuring that the semantic integrity of the text remains intact while enhancing the model's ability to memorize the watermarked data.
Through extensive experimentation on models such as LLaMA-3 8B and the \textit{BookMIA} and \textit{The Pile} datasets, \methodname ~demonstrated high improvements in AUROC scores for MIA detection, consistently exceeding 90\% on the BookMIA dataset across multiple detection methods and models, highlighting its reliability in identifying watermarked data.
We also evaluated the semantic preservation of the watermarked text and explored various synonym substitution methods to identify the optimal approach that balances semantic integrity with high detection accuracy. This offers a practical solution for organizations aiming to protect their datasets without compromising usability or content clarity.

Future work will focus on refining the synonym substitution method to further optimize the balance between watermark detectability and model memorization, ensuring that the method maintains high watermark detection rates without impacting model performance. 
Another key focus will be examining the effects of watermarking during pre-training by assessing how the technique influences model learning dynamics when trained on watermarked datasets.
Expanding our approach to larger models and multilingual datasets will also be a priority, addressing the need for versatile watermarking solutions across a broader range of applications. 
While our current implementation uses English-centric tools, \methodname\ is inherently language-agnostic. Resources such as Open Multilingual WordNet~\cite{bond-foster-2013-linking} and multilingual BERT (mBERT)~\cite{devlin2019bertpretrainingdeepbidirectional} can support synonym substitution in many languages, including low-resource ones. 



By continuing to advance watermarking methods, \methodname ~provides a scalable and practical solution for addressing the critical need to monitor and detect the use of proprietary data in LLM training. By enhancing dataset traceability, \methodname ~serves as a practical tool, ultimately contributing to the ethical and secure development of LLM applications.\\






\bibliographystyle{IEEEtran}
\bibliography{sample}

\begin{appendices}

\section{Analysis of Top-K Value Selection}
\label{app:k_values}
\begin{table}[ht]
\centering
\caption{Detailed comparison of AUROC, semantic preservation, and BLEU scores across different k-values using the MIA method: MinK++\_20.0\% and synonym replacement method: BERT th=0.8.}
\small
\begin{tabular}{cccc}
\toprule
\textbf{K} & \textbf{AUROC} & \textbf{CosSim} & \textbf{BLEU} \\
\midrule
\multirow{1}{*}{3} & 87.5\% & \textbf{98.4} & \textbf{0.75} \\
\midrule
\multirow{1}{*}{4} & 91.8\% & 94.22 & 0.67 \\
\midrule
\multirow{1}{*}{5} &\textit{ 94.0\%} & \textit{88.33} & \textit{0.6} \\
\midrule
\multirow{1}{*}{6} & 95.6\% & 81.72 & 0.55 \\
\midrule
\multirow{1}{*}{7} & \textbf{96.6\%} & 76.43 & 0.5 \\ 
\bottomrule
\end{tabular}
\label{tab:comparison_k_values}
\end{table}

Table~\ref{tab:comparison_k_values} presents the results of using different K values in selecting the top-K words in each sentence. Reported are the AUROC score, using the MIA method of \textit{MinK++ 20.0\%}, compared to the semantic similarity between the watermarked text and the original text, measured by the cosine similarity of embeddings with a threshold of 0.8, as well as the BLEU score. For synonym replacement, we used a BERT model with a threshold of 0.8.

The findings reveal a trade-off between the AUROC and semantic preservation: lower K values, which introduce fewer changes to the text, tend to maintain higher semantic similarity. However, increasing K improves detection accuracy at the cost of reduced semantic preservation.
In our experiments, we selected K=5 maintain a balance between detection accuracy and semantic similarity. While higher K values provide only marginal improvements in the AUROC, they highly impact semantic integrity, making K=5 a suitable compromise.



\section{Efficiency of Synonym Retrieval Methods}
\label{app:synonym_runtime}
A potential concern with our approach is the computational overhead introduced by embedding-based synonym generation, particularly when using large language models such as BERT. To assess the practical implications, we benchmarked several synonym retrieval methods on the Enron Email dataset. The methods evaluated include SBERT, context-based BERT, two variants of our LexSub method (concatenation and dropout), and a lightweight alternative based on WordNet. The average runtime per email and total processing time are summarized in Table~\ref{tab:synonym_runtime}.

\begin{table}[h]
\centering
\caption{Average runtime per email for different synonym retrieval methods. Experiments conducted on an NVIDIA RTX 6000 GPU.}
\label{tab:synonym_runtime}
\begin{tabular}{lc}
\toprule
\textbf{Method} & \textbf{Average Time per Email (sec)} \\
\midrule
WordNet        & 0.6211 \\
SBERT          & 2.0153 \\
Concatenation  & 2.1300 \\
Context BERT   & 2.7154 \\
Dropout        & 4.8400 \\
\bottomrule
\end{tabular}
\end{table}

As shown, WordNet offers a highly efficient option that does not rely on model inference, making it suitable for large-scale or resource-constrained applications. SBERT and the LexSub concatenation method also exhibit relatively low latency, averaging around 2 seconds per document. While the LexSub variant with dropout introduces more computational overhead, it remains practical for real-world deployment.

These findings demonstrate that LexiMark supports efficient and flexible deployment. Depending on the available computational resources and desired fidelity of contextual understanding, users can choose between fast dictionary-based methods or more sophisticated neural approaches.


\end{appendices}

\end{document}